\documentclass{article}
\usepackage{babel}
\usepackage{arxiv}
\usepackage{graphicx}
\usepackage[T1]{fontenc}
\usepackage{amsthm}

\usepackage{amsmath}
\usepackage{amssymb}
\usepackage{booktabs}
\usepackage{amsfonts}
\usepackage{nicefrac}
\usepackage{microtype}
\usepackage{cleveref}

\usepackage{algorithm}
\usepackage{algpseudocode}

\newtheorem{theorem}{Theorem}
\newtheorem{lemma}{Lemma}

\newtheorem{statement}{Statement}
\newtheorem{definition}{Definition}
\newtheorem*{remark}{Remark}
\newtheorem*{corollary}{Corollary}

\author{Maria Nikitina \and  Anton Bishuk \and Oleg Bakhteev}

\title{Spectral Characteristics of Autoencoder Parameters as a Vector Representation of Data}%
\renewcommand{\headeright}{}
\renewcommand{\undertitle}{}
\date{}
\begin{document}

\maketitle
\begin{abstract}\ \footnote{This is a shortened (theorem proofs are skipped) and translated version of the paper published in a Russian-language peer-reviewed journal. For citation, please use the following reference: Bishuk A., Nikitina M., Bakhteev O. ``Spectral characteristics of autoencoder parameters as a vector representation of data.'' Upravlenie bolsimi sistemami. 2026. Vol. 120. P. 51–83.}
This paper examines the relationship between the parameters of autoencoder models and the statistical properties of the data on which they are trained. Autoencoders are defined as models with an encoder-decoder architecture, trained to reconstruct input data through a compressed latent representation. It is proposed that the model parameters can be viewed as a dense vector representation of the corresponding sample. To test this hypothesis, a theoretical and experimental study is conducted in which a vector representation is formed based on the spectral characteristics of the autoencoder parameter matrices. Theoretical analysis shows that the singular values of the model parameter matrices are related to the eigenvalues of the covariance matrix of the training data, ensuring the transfer of information between the data space and the parameter space. Experimental results on the CIFAR-10 and FashionMNIST datasets confirm that the resulting vector representations allow for a high degree of accuracy in distinguishing between models trained on different data subsets, without resorting to complex vector generation algorithms or using the original samples. These results suggest that the parameters of trained autoencoders can be viewed as sample representations.
\end{abstract}

\
\keywords{machine learning \and neural networks \and vector representation \and autoencoder \and spectral characteristics \and generative models \and embeddings}

\section{Introduction}
Machine learning methods based on artificial neural networks have become firmly established as an important tool applied in various areas of modern data analysis~\cite{Lagovskiy2023, Saraev2010}. The relationship between the structure of trained neural networks and the statistical properties of the data on which they were trained is one of the fundamental problems of modern machine learning~\cite{Model_and_data}. Understanding how the distribution of training data is reflected in the model parameters is of key importance for such areas as neural network interpretability, analysis of data properties, architecture selection, and the construction of meta-representations of models~\cite{Achille2019Information, Adilova2023, Arpit2017Generalization, Neyshabur2017Implicit, Zhang2021}. Furthermore, this problem is of particular relevance to generative models, since they not only extract patterns but also form internal representations that reflect the structure of the data distribution~\cite{Yang2022, Rezende2014Stochastic}. This paper focuses on autoencoders, neural networks based on the encoder--decoder architecture. Although their classical architecture is not formally generative, regularized variants approximate or even constitute full-fledged generative models~\cite{alain2014regularized, Kingma2022VAE, Refinetti2023LAE}. Thus, the analysis of the parameter structure of an autoencoder is naturally connected to the broader class of generative models, and the results of this work can potentially be generalized to this class.

A broad range of studies has been devoted to the construction of embeddings of machine learning models. Classical approaches rely on encoding architectural parameters~\cite{Cheng2024, Elsken2019NAS}, meta-representations obtained using separate models~\cite{Cui2025, hyperrepr_generative}, or representations constructed from the model's outputs on test data~\cite{Dataset2Vec}. Recent works have proposed treating a model as a point in a continuous space, enabling model search, interpolation, and analysis~\cite{Alet2020Modular, Ilharco2022Editing}. However, most of these methods either require direct access to the original data or rely on complex procedural heuristics that combine intermediate computations, architectural features, or meta-information.

In this paper, a different approach is considered. We investigate the extent to which the parameters of a trained model can serve as a representative description of the data on which it was trained. Specifically, we test the hypothesis that the spectral characteristics of the parameters of a trained neural network form a compact embedding of the original dataset. Thus, the trained model acts as an embedding of the dataset, while its parameters serve as a carrier of the statistical characteristics of the underlying data distribution~\cite{Klabunde2025Similarity, Raghu2017SVCCA}.

To demonstrate the proposed approach, we introduce a model vectorization method based on the analysis of the singular values of the model parameter matrices. It is shown that the proposed method enables the separation of models trained on different datasets with high accuracy. This result indicates the existence of a deep relationship between the spectral properties of model parameters and the statistics of the training datasets. In contrast to approaches based on external features or approximations of the data distribution, the proposed method operates exclusively in the parameter space of the model, without access to the original data or additional training of auxiliary networks.

The theoretical part of the paper provides a justification for the relationship between the singular values of the model parameters and the spectrum of the covariance matrix of the data. It is shown that, for autoencoders with fully connected layers, the singular values of the trained weight matrices contain information about the distribution of the training datasets. Furthermore, it is proven that if the underlying datasets differ, the resulting embeddings also differ.

The experimental part is aimed at testing the hypothesis that even when using shallow architectures from the families of autoencoders (AE) and variational autoencoders (VAE), together with non-trainable vectorization methods based solely on model parameters, it is possible to distinguish models trained on different subsets of data with high accuracy. Experiments on the CIFAR-10~\cite{CIFAR10} and FashionMNIST~\cite{FashionMNIST} datasets confirm that classifying such models using their embeddings achieves high values of standard classification metrics, including Accuracy, ROC AUC, and Average Precision. Moreover, the gradual addition of new classes or noise to the training dataset is reflected in the resulting embeddings in a predictable manner, indicating that the model space possesses a meaningful geometric structure.

The obtained results make it possible to consider the parameters of neural networks as a means of describing data. In this sense, trained models become not only a tool for data generation or approximation but also a natural embedding for analyzing the structure of datasets and the relationships between them.


\section{Problem statement}
Let $p_c$ denote the data distribution of class $c \in \{1,2,\ldots\}$ over the space $\mathbb{R}^d$ of $d$-dimensional vectors. Let the dataset be given by $\mathbf{X}_c = \{\mathbf{x}_i : \mathbf{x}_i \sim p_c\}_{i=1}^{n}$. A dataset may consist of several classes, each corresponding to its own distribution $p_c$. In the classical unsupervised learning setting, $c \equiv 1$ and $p_c \equiv p_1 \equiv p$, while $\mathbf{X}_c \equiv \mathbf{X}_1 \equiv \mathbf{X}$.

We consider a family of autoencoder models. A model is defined as a parameterized function $\mathbf{f}_{\boldsymbol{\psi}} : \mathbb{R}^d \to \mathbb{R}^d$, represented as the composition of an encoder and a decoder, $\mathbf{f}_{\boldsymbol{\psi}} = D_{\boldsymbol{\varphi}} \circ E_{\boldsymbol{\theta}}$. The model parameters are denoted by $\boldsymbol{\psi} = (\boldsymbol{\theta}, \boldsymbol{\varphi}) \in \Psi$, where $E_{\boldsymbol{\theta}} : \mathbb{R}^d \to \mathbb{R}^z$ is the encoder with parameters $\boldsymbol{\theta}$, and $D_{\boldsymbol{\varphi}} : \mathbb{R}^z \to \mathbb{R}^d$ is the decoder with parameters $\boldsymbol{\varphi}$. Here, $z$ denotes the dimension of the hidden space. Training is performed by minimizing the objective
\[
\mathcal{L}(\boldsymbol{\psi};p_c)
:= -\mathbb{E}_{\mathbf{x}\sim p_c}
\Bigl[
l_{\boldsymbol{\psi}}(\mathbf{x})
\Bigr],
\]
where the maximized functional $l_{\boldsymbol{\psi}}(\mathbf{x})$ is defined as
\[
l_{\boldsymbol{\psi}}(\mathbf{x})
=
\begin{cases}
-\ell\bigl(\mathbf{x},\, \mathbf{f}_{\boldsymbol{\psi}}(\mathbf{x})\bigr),
& \text{for AE},\\[6pt]
\begin{aligned}[t]
&\mathbb{E}_{\mathbf{h}\sim q_{\boldsymbol{\theta}}(\mathbf{h}\mid \mathbf{x})}
\bigl[-\ell(\mathbf{x}, D_{\boldsymbol{\varphi}}(\mathbf{h}))\bigr] \\
&\qquad- \mathrm{KL}\!\left(q_{\boldsymbol{\theta}}(\mathbf{h}\mid \mathbf{x})\,\|\,p(\mathbf{h})\right),
\end{aligned}
& \text{for VAE},
\end{cases}
\]
where $\ell: \mathbb{R}^d \times \mathbb{R}^d \to \mathbb{R}$ is the reconstruction loss function (e.g., mean squared error or cross-entropy), which measures the discrepancy between the input vector $\mathbf{x}$ and its reconstruction; $\mathrm{KL}(\cdot \| \cdot)$ denotes the Kullback--Leibler divergence, a measure of the difference between two probability distributions; $q_{\boldsymbol{\theta}}(\mathbf{h}\mid \mathbf{x})$ is the approximate (variational) distribution in the latent space parameterized by the encoder; and $p(\mathbf{h})$ is the prior distribution in the latent space (typically the multivariate normal distribution $\mathcal{N}(\mathbf{0}, \mathbf{I})$).

For a standard autoencoder (AE), the objective reduces to minimizing the reconstruction error. For a variational autoencoder (VAE), the objective consists of two terms: the expected reconstruction error (where the expectation is taken over the latent variable $\mathbf{h}$) and the KL regularization term, which encourages the encoder distribution $q_{\boldsymbol{\theta}}(\mathbf{h}\mid \mathbf{x})$ to be close to the prescribed prior distribution $p(\mathbf{h})$.

Suppose that a set of generative models is available, each trained on its own dataset representing a particular underlying population. The goal is to construct a representation space in which each model is mapped to a dense embedding vector preserving the statistical properties of the original data. Such a space should enable the analysis of both the datasets and the models trained on them without requiring retraining or direct access to the original data. At the same time, the resulting embedding is not required to uniquely reconstruct the original model but only to describe a predefined set of its properties.

The key property of the desired representation is the preservation of the ability to discriminate models trained on datasets generated from different distributions $p$. Suppose that there exists a representation space in which the datasets $\mathbf{X}_{a} \in \mathbb{R}^{n_{a}\times d}$, $\mathbf{X}_{b} \in \mathbb{R}^{n_{b}\times d}$, and $\mathbf{X}_{c} \in \mathbb{R}^{n_{c}\times d}$ are linearly separable. The same property is required for the embeddings of generative models trained on subsets of samples from these datasets.

We introduce embeddings of both models and datasets based on singular values. Let the model parameters $\boldsymbol{\psi}$ be represented as the set of weight matrices $\{\mathbf{W}_l\}_{l=1}^L$ corresponding to its layers.

The model singular values $\boldsymbol{\sigma}^{\text{model}}(\boldsymbol{\psi})$ are defined algorithmically (see Algorithm~\ref{alg:1}): for each weight matrix $\mathbf{W}_l$, the $r$ largest singular values are computed; the resulting embeddings are concatenated in the order of the layers and normalized. Thus, the vector $\boldsymbol{\sigma}^{\text{model}}(\boldsymbol{\psi})$ is uniquely determined by the parameters and the architecture of the model. The data singular values $\boldsymbol{\sigma}^{\text{data}}(\mathbf{X}_c)$ for a dataset $\mathbf{X}_c$ are defined as the vector of its $m$ largest singular values sorted in descending order.

\begin{definition}
Let the sets of vectors ${\mathbf{X}_a \in \mathbb{R}^{n \times d}}$ and ${\mathbf{X}_b \in \mathbb{R}^{m \times d}}$ be given. Define mixing as the concatenation of vectors from $\mathbf{X}_a$ and $\mathbf{X}_b$ to obtain ${\tilde{\mathbf{X}}_{a|b} \in \mathbb{R}^{(n + m) \times d}}$.

$\alpha$-mixing is defined as the concatenation of $\mathbf{X}_a$ and ${\lfloor\alpha m\rfloor, ~ \alpha \in [0; 1]}$ vectors from $\mathbf{X}_b$ to obtain ${\tilde{\mathbf{X}}_{a|b} \in \mathbb{R}^{(n + \lfloor\alpha m\rfloor) \times d}}$.
\end{definition}

\begin{figure}[ht]
    \centering
    \includegraphics[width=0.8\linewidth]{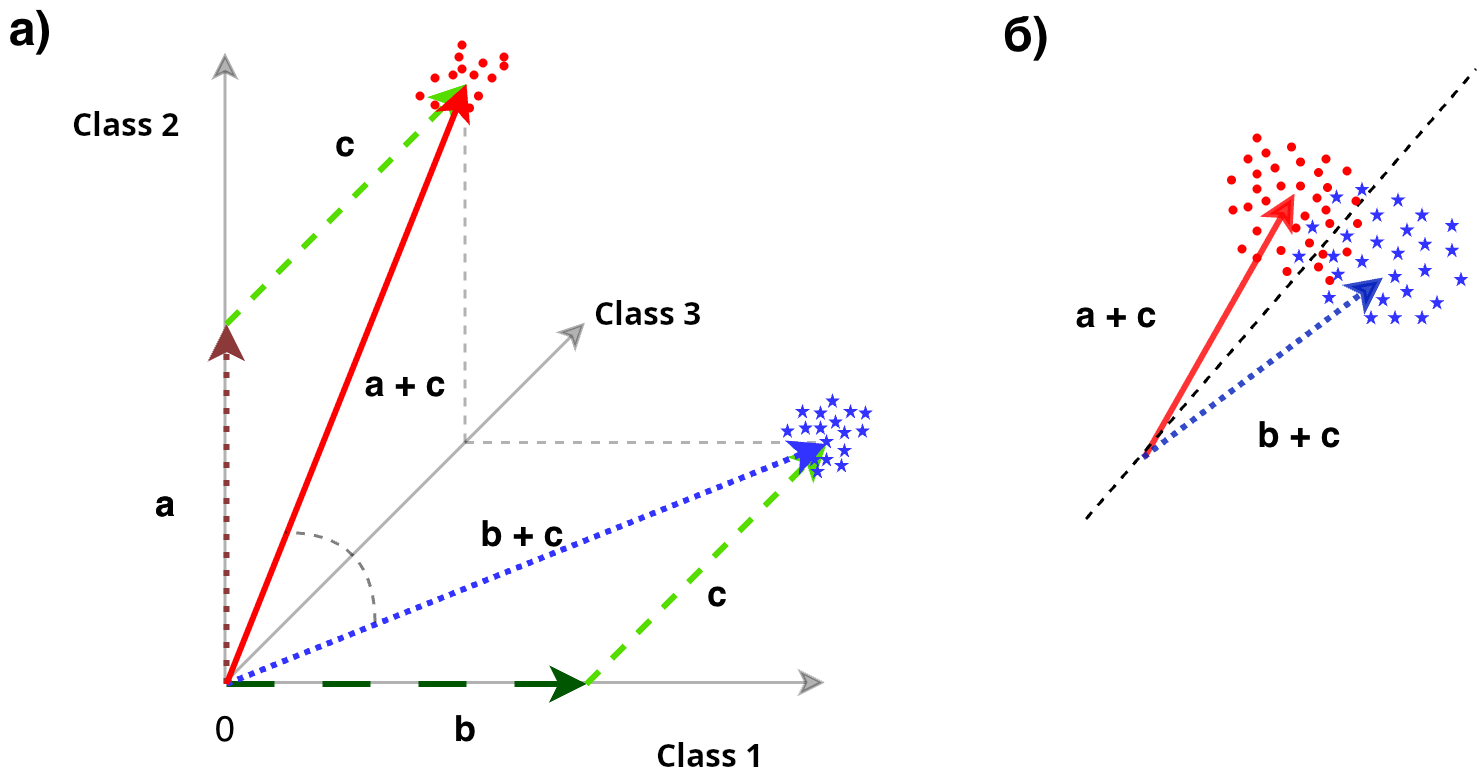}
    \caption{Schematic representation of the relationships between datasets in the embedding space: \textbf{a)} When vectors from a third dataset $\mathbf{c}$ are mixed into samples $\mathbf{a}$ and $\mathbf{b}$, the embeddings of the resulting samples $\mathbf{a} + \mathbf{c}$ and $\mathbf{b} + \mathbf{c}$ become closer; \textbf{b)} When the vectors of the resulting samples become sufficiently close, they cease to be linearly separable.}
    \label{fg:classes}
\end{figure}

Then, mixing samples from $\mathbf{X}_{c}$ into $\mathbf{X}_{a}$ and $\mathbf{X}_{b}$ should bring the resulting datasets ${\mathbf{X}_{a + c}, \mathbf{X}_{b + c}}$ closer to each other, as illustrated in Fig.~\ref{fg:classes} (a), i.e., reduce their linear separability, as shown in Fig.~\ref{fg:classes} (b). The embeddings of generative models trained on samples from the selected datasets should exhibit the same properties.

Furthermore, the representation should be robust to variations within the same data distribution while remaining sensitive to significant statistical differences between datasets. An important task is to assess how well the model embeddings preserves information about the original dataset. To this end, it is necessary to consider not only the accuracy of reconstructing the original data but also the ability of the representation to reflect the structure and relationships between datasets.

Thus, the task reduces to constructing a mapping from the parameters of a generative model to a compact embedding space while preserving the statistical properties of the original data and their combinations. To address this task, we use a method based on the spectral characteristics of the model parameter matrices, which enables the construction of interpretable and informative embeddings.


\section{Proposed method}
The proposed method is based on the idea of representing a model $f$ as a dense embedding containing information about the statistical properties of the data $X_c$ on which it was trained. For this purpose, we use $\boldsymbol{\sigma}^{\text{model}}(\boldsymbol{\psi})$ --- the singular values of the parameter matrices $\boldsymbol{\psi}$ extracted from the trained model. This choice is motivated by the fact that the spectral structure of neural network parameters reflects key properties of the input data distribution, while $\boldsymbol{\sigma}^{\text{model}}$ is sensitive to statistical differences between datasets, as will be theoretically justified below.

The formalized scheme of Algorithm~\ref{alg:1} illustrates the key steps of the proposed approach.

\begin{algorithm}[ht]
\caption{Construction of embeddings for a generative model}
\label{alg:1}
\begin{algorithmic}[1]
\Require A trained model with $L$ layers, parameter matrices $\{\mathbf{W}_l\}_{l=1}^L$. $r$ --- the number of singular values selected to construct the model embedding.
\Ensure Embeddings of the model $\mathbf{v}$
\State Initialize an empty feature vector $\mathbf{v} \gets []$;
\For{each layer $\mathbf{W}_l$ in the model}
    \State Perform the singular value decomposition $\mathbf{W}_l = \mathbf{U}_l \mathbf{S}_l \mathbf{V}_l^T$;
    \State Extract the singular values $\sigma_{l,1}^{\text{model}}, \dots, \sigma_{l,k_l}^{\text{model}}$ from $\mathbf{S}_l$;
    \State Sort the singular values in descending order;
    \State Keep the $r$ largest singular values;
    \State Add the resulting values to $\mathbf{v}$.
\EndFor
\State Normalize $\mathbf{v}$ (e.g., to unit norm);
\State \Return $\mathbf{v}$
\end{algorithmic}
\end{algorithm}

\begin{remark}
The number $r$ in the algorithm can be chosen as the minimum number of singular values of the first layer that account for a predefined fraction of the sum of all singular values of that layer.
\end{remark}

To theoretically justify the proposed method, it is necessary to assess the relationship between $\boldsymbol{\sigma}^{\text{model}}$ and $\boldsymbol{\sigma}^{\text{data}}$. Suppose that there exists an embedding space in which the singular value vectors of different datasets are linearly separable. We aim to prove that the singular value vectors of the parameters of generative models trained on subsets of these datasets can also be linearly separated.

\begin{statement}
\label{statement}
The closer the singular value vectors of the original datasets are, the worse the linear separability of the singular value vectors of the corresponding models.
\end{statement}

Let ${V_{\sigma} \subset \mathbb{R}^d}$ be the space of singular value vectors of datasets. To prove hypothesis~\ref{statement}, the convergence of ${\boldsymbol{\sigma}^{\text{data}}(\mathbf{X}_1) \in V_{\sigma}}$ and ${\boldsymbol{\sigma}^{\text{data}}(\mathbf{X}_2) \in V_{\sigma}}$ for the data matrices $\mathbf{X}_1$ and $\mathbf{X}_2$, respectively, is achieved by mixing samples from a third dataset $\mathbf{X}_3$. To demonstrate that the singular values of $\tilde{\mathbf{X}}_{c|3}$, obtained by concatenating $\mathbf{X}_c$ and $\mathbf{X}_3$ for ${c \in \{1, 2\}}$, converge as the fraction of samples from $\mathbf{X}_3$ increases, Lemma~\ref{lemma:1} was proven. For notational simplicity, below we use ${\tilde{\mathbf{X}}_{c} \equiv \tilde{\mathbf{X}}_{c|3}}$ for ${c \in \{1, 2\}}$.

\begin{lemma}\label{lemma:1}
Let ${\mathbf{X}_1 \in \mathbb{R}^{n_1 \times d}}$ be the data matrix of the first dataset ($n_1$ samples, $d$ features), ${\mathbf{X}_2 \in \mathbb{R}^{n_2 \times d}}$ be the data matrix of the second dataset, and ${\mathbf{X}_3 \in \mathbb{R}^{m \times d}}$ be the data matrix of the third dataset.

Then $\boldsymbol{\sigma}^{\text{data}}(\tilde{\mathbf{X}}_1)$ and $\boldsymbol{\sigma}^{\text{data}}(\tilde{\mathbf{X}}_2)$ --- the singular values of the matrices $\tilde{\mathbf{X}}_1$ and $\tilde{\mathbf{X}}_2$ --- converge:
${|\sigma_i^{\text{data}}(\tilde{\mathbf{X}}_1) - \sigma_i^{\text{data}}(\tilde{\mathbf{X}}_2)| \longrightarrow 0}$ as $m \to \infty$, where $\sigma_i^{\text{data}}(\tilde{\mathbf{X}}_c)$ is the $i$-th singular value of the dataset $\tilde{\mathbf{X}}_c$ obtained by mixing vectors from $\mathbf{X}_3$ into $\mathbf{X}_c$.
\end{lemma}

In other words, the singular values of the mixtures $\tilde{\mathbf{X}}_c$ converge as the fraction of samples from $\mathbf{X}_3$ increases. We can now proceed to the proof of Theorem~\ref{theorem:1} concerning the relationship between the singular values of the dataset and the model weights.

The theorem is formulated and proved for a special case of an autoencoder: an autoencoder with a single fully connected layer in both the encoder and decoder, without a nonlinear activation function. We introduce $\mathbf{W}_{c}^{enc}$ as the matrix representation of the encoder parameters $\boldsymbol{\theta}_c$, and $\mathbf{W}_{c}^{dec}$ as the matrix representation of the decoder parameters $\boldsymbol{\varphi}_c$.

\begin{theorem}\label{theorem:1}
Let three datasets be given: ${\mathbf{X}_1 \in \mathbb{R}^{n \times d}}$, ${\mathbf{X}_2 \in \mathbb{R}^{n \times d}}$, and ${\mathbf{X}_3 \in \mathbb{R}^{m \times d}}$; let their singular values be positive: ${\sigma_i^{\text{data}}(\mathbf{X}_c) > 0 ~~~ \forall i \in \overline{1, d}}$.

Let two new datasets be obtained by mixing vectors from the third dataset into each of the first two datasets: ${\tilde{\mathbf{X}}_c = [\mathbf{X}_c, \mathbf{X}_3] \in \mathbb{R}^{(n+m) \times d}, ~~~ c \in \{1, 2\}}$.

Suppose that an autoencoder model with fully connected layers without nonlinear activation functions and latent dimension $k$ is trained on each dataset, minimizing the loss function
${\min_{\boldsymbol{\psi}_c}(\mathcal{L}(\boldsymbol{\psi}_c)) = \min_{\boldsymbol{\psi}_c} \|\tilde{\mathbf{X}}_c - \tilde{\mathbf{X}}_c\mathbf{W}_{c}^{enc}\mathbf{W}_{c}^{dec\top}\|_F^2}$
with a regularization term of the form ${\|\text{cov}(\mathbf{h}) - \mathbf{I}\|^2}$, which decorrelates the vectors $\mathbf{h}$ in the latent space, i.e., ${\text{cov}(\mathbf{h}) \sim I}$.

Then the mean squared difference between the singular values of the parameters of the autoencoders with parameters $\mathbf{W}^{enc}_1$, $\mathbf{W}^{dec}_1$, $\mathbf{W}^{enc}_2$, $\mathbf{W}^{dec}_2$ decreases as $m$ increases:
\[
\|\sigma_i^{\text{model}}(\mathbf{W}^{enc}_1) - \sigma_i^{\text{model}}(\mathbf{W}^{enc}_2)\|_2 \underset{m \to \infty}{\longrightarrow} 0,
\]
Consequently, as the datasets converge, both their singular values and the singular values of the parameters of the autoencoders trained on these datasets converge as the number of rows $m$ tends to infinity.
\end{theorem}

\begin{corollary}
The loss function $\mathcal{L}(\boldsymbol{\psi}_c)$ with the addition of the regularization term ${\|\text{cov}(\mathbf{h}_c) - \mathbf{I}\|^2}$, which decorrelates the vectors $\mathbf{h}_c$ in the hidden space, makes the autoencoder closer in its properties to a VAE, since the covariance matrix of the hidden (latent) representation becomes diagonal, which is characteristic of a VAE. This observation allows the results of Theorem~\ref{theorem:1} to be used as evidence that hypothesis~\ref{statement} holds for both AE and VAE models.
\end{corollary}

\begin{remark}
Theorem~\ref{theorem:1} is proved for the special case of models with fully connected layers without nonlinearities. The applicability of the theorem to the nonlinear case is evaluated experimentally. A theorem with a larger number of restrictions on the problem being solved, but including models with nonlinearities, is presented below.
\end{remark}

Theorem~\ref{theorem:1} characterizes the behaviour of autoencoder embeddings as the vectors of the datasets converge. However, the greater practical value lies in the ability of models to describe not similarities but differences between datasets. As a basic property, we consider the ability to distinguish models trained on datasets generated from different distributions.

Let $p_c$ denote the data distribution of class ${c\in\{1,2\}}$ on $\mathbb{R}^d$ with covariance ${\boldsymbol{\Sigma}_c := \boldsymbol{\Sigma}_c(p_c) := \mathbb{E}_{\mathbf{x}\sim p_c}[\mathbf{x}\mathbf{x}^\top]}$. For a dataset ${\mathbf{X}_c=\{\mathbf{x}_i: \mathbf{x}_i \sim p_c\}_{i=1}^n}$ of size $n$, the empirical covariance is defined as ${\widehat{\boldsymbol{\Sigma}}_c := \widehat{\boldsymbol{\Sigma}}_c (\mathbf{X}_c) := \frac{1}{n}\sum_{i=1}^n \mathbf{x}_i \mathbf{x}_i^\top}$.

We consider a model with parameters ${\boldsymbol{\psi}\in\Psi}$ and loss function ${\mathcal{L}(\boldsymbol{\psi};p_c):= -\mathbb{E}_{\mathbf{x}\sim p_c}\Bigl[l_{\boldsymbol{\psi}}(\mathbf{x})\Bigr]}$, where $l_{\boldsymbol{\psi}}(\mathbf{x})$ is some maximized functional.

In this work, we assume that the loss function $\mathcal{L}(\boldsymbol{\psi};p_c)$ can be expressed as a function of the data covariance $\boldsymbol{\Sigma}_c$, i.e., there exists a functional $\mathcal{L}_{\boldsymbol{\Sigma}}$ such that ${\mathcal{L}(\boldsymbol{\psi};p_c) = \mathcal{L}_{\boldsymbol{\Sigma}}(\boldsymbol{\psi};\boldsymbol{\Sigma}_c(p_c))}$. This holds, for example, when $l_{\boldsymbol{\psi}}(\mathbf{x})$ depends on $\mathbf{x}$ only through its second moments, or when the distribution $p_c$ is completely determined by its covariance, as in the Gaussian case.

For notational simplicity, we use the notation ${\mathcal{L}(\boldsymbol{\psi};\boldsymbol{\Sigma}_c) := \mathcal{L}_{\boldsymbol{\Sigma}}(\boldsymbol{\psi};\boldsymbol{\Sigma}_c)}$ below.

Suppose that $\mathcal{L}(\boldsymbol{\psi};\boldsymbol{\Sigma}_c)$ attains its minimum at $\boldsymbol{\psi}^*(\cdot)$. On the dataset $\mathbf{X}_c$, the minimum is attained at $\widehat{\boldsymbol{\psi}}_n(\cdot)$:
\[
\boldsymbol{\psi}^*(\boldsymbol{\Sigma}_c):=\arg\min_{\boldsymbol{\psi}\in\Psi} \mathcal{L}(\boldsymbol{\psi};\boldsymbol{\Sigma}_c), ~~~ \widehat{\boldsymbol{\psi}}_n(\widehat{\boldsymbol{\Sigma}}_c):=\arg\min_{\boldsymbol{\psi}\in\Psi} \mathcal{L}_n(\boldsymbol{\psi};\widehat{\boldsymbol{\Sigma}}_c),
\]
\noindent where $\mathcal{L}_n$ is the empirical loss function.

For a proper formulation and proof of the subsequent results, it is necessary to introduce a set of assumptions ensuring the smoothness, convexity, and stability of the optimal solutions, as well as the boundedness of the data.

\textbf{Theorem conditions.}
\begin{enumerate}
    \item[(B1)] \textbf{(Smoothness)} \(\mathcal{L}(\boldsymbol{\psi};\boldsymbol{\Sigma}_c)\) belongs to the class \(C^2\) with respect to \(\boldsymbol{\psi}\) and to \(C^1\) with respect to the covariance parameter \(\boldsymbol{\Sigma}_c\). Note that verifying smoothness with respect to $\boldsymbol{\Sigma}_c$ is computationally difficult. However, the statement holds for Gaussian distributions~\cite{drton2009smoothness}.
    
    \item[(B2)] \textbf{(Local uniqueness and strong convexity)} For each class \(c\), there exists a neighborhood \(U_c^{B2}\) of covariance matrices around \(\boldsymbol{\Sigma}_c\), defined with respect to the metric ${\|\cdot\|_{op}}$, such that for all \({\boldsymbol{\Sigma}\in U_c^{B2}}\), the function \({\boldsymbol{\psi}\mapsto\mathcal{L}(\boldsymbol{\psi};\boldsymbol{\Sigma}_c)}\) has a locally unique isolated minimum \(\boldsymbol{\psi}^*(\boldsymbol{\Sigma})\), and the Hessian at this point satisfies
    \[
    {\exists \hat{\mu}: \lambda_{\min}\big(\nabla^2_{\boldsymbol{\psi}\boldsymbol{\psi}}\mathcal{L}(\boldsymbol{\psi}^*(\boldsymbol{\Sigma});\boldsymbol{\Sigma}_c)\big)\ge\hat{\mu}>0,}
    \]
    where $\lambda_{\min}(\cdot)$ denotes the smallest eigenvalue of the Hessian.

    \item[(B3)] \textbf{(Differentiability with respect to \(\boldsymbol{\Sigma}\).)}
    For each class \(c\), there exists a neighborhood \(U_c^{B3}\) of covariance matrices around \(\boldsymbol{\Sigma}_c\), in which the family of minimizers \(\boldsymbol{\psi}^*(\boldsymbol{\Sigma})\) is a \(C^1\)-function of \(\boldsymbol{\Sigma}\), i.e., the Jacobian
    ${J_{\boldsymbol{\Sigma}} := \partial_{\boldsymbol{\Sigma}}\boldsymbol{\psi}^*(\boldsymbol{\Sigma})}$
    exists and is continuous.

    For each class, we consider the neighborhood ${U_c = U_c^{B2} \cap U_c^{B3}}$, so that both requirements (B2) and (B3) hold simultaneously. Since these are neighborhoods of the same point $\boldsymbol{\Sigma}_c$, their intersection is non-empty.

    \item[(B4)] \textbf{(Approximation error)}
    There exists a fixed neighborhood ${\mathcal V_c\subset\Psi}$ and a sequence ${\eta_n^c \overset{p}\to 0}$ such that
    ${\sup_{\boldsymbol{\psi}\in\mathcal V_c}\big|\mathcal{L}_n(\boldsymbol{\psi};\widehat{\boldsymbol{\Sigma}}_c)-\mathcal{L}(\boldsymbol{\psi};\boldsymbol{\Sigma}_c)\big| \leq \eta_n^c}$, where $n$ is the sample size. All values of $\boldsymbol{\psi}^*(\boldsymbol{\Sigma}_c)$ considered in the theorem lie in $\mathcal V_c$. To generalize to all classes $c$ in the theorem, we use the sequence obtained by taking the pointwise maximum ${\eta_n := \max(\eta_n^1, \eta_n^2, \ldots \eta_n^c)}$.
    
    \item[(B5)] \textbf{(Bounded data)}
    We assume that there exists a constant $B > 0$ such that ${\|\mathbf{x}_i\|_2 \leq B}$ almost surely for all $i$. This property ensures the applicability of concentration inequalities to the empirical covariance, and in practice can be achieved by preprocessing the data using clipping or winsorization.

    \item[(B6)] \textbf{(Distinct first $r$ singular values)}
    Suppose that the first $r$ singular values of the parameter matrix $\boldsymbol{\psi}^*(\boldsymbol{\Sigma}_c)$ are distinct and separated from the remaining singular values by ${\delta_r > 0}$.
\end{enumerate}

\begin{theorem}\label{thm:vae_singulars}
Suppose that assumptions (B1)--(B6) hold. We introduce the notation ${\Delta_r := \|\boldsymbol{\sigma}_{1:r}^{data}(\mathbf{X}_1)-\boldsymbol{\sigma}_{1:r}^{data}(\mathbf{X}_2)\|_2,}$ where $\boldsymbol{\sigma}^{data}_{1:r}(\cdot)$ denotes the singular values of the data matrices $\mathbf{X}_c$ from the first to the $r$-th.

\noindent Then there exist constants \({C_0>0}\) and \({c_*>0}\) such that, for any \({\delta\in(0;\frac{1}{2})}\), for a sample size \(n\), with probability at least \({1-2\delta}\),
\[
\begin{aligned}
\big\|\boldsymbol{\sigma}_{1:r}^{model}(\widehat{\boldsymbol{\psi}}_n(\widehat{\boldsymbol{\Sigma}}_1))-
\boldsymbol{\sigma}_{1:r}^{model}(\widehat{\boldsymbol{\psi}}_n(\widehat{\boldsymbol{\Sigma}}_2))\big\|_2
\ge c_*\Delta_r - C_0\Bigl(\sqrt{\tfrac{\log(1/\delta)}{n^2}} + \sqrt{\eta_n} + \tau_n\Bigr),
\end{aligned}
\]
where \({\tau_n := \max(\tau_n^1, \tau_n^2)}\) is the maximum of the optimization errors for models trained on samples of size $n$ using $\mathbf{X}_1$ and $\mathbf{X}_2$.

In particular, if \({\Delta_r>0}\) and \({\eta_n,\tau_n \to 0}\), then as \({n \to \infty}\), the right-hand side remains positive, and the vectors of the first \(r\) singular values remain separated by a nonzero gap with probability tending to 1.
\end{theorem}

This result demonstrates that the embedding of a generative model contains sufficient information about the distribution of the original data. Moreover, such representations may possess a structure that reflects the mixing of datasets and the possibility of linearly separating them.

\begin{corollary}
In the case of a VAE model, the functional $l_{\boldsymbol{\psi}}(\mathbf{x})$ is the $\mathrm{ELBO}$, which possesses all the necessary properties for the conditions of the theorem to hold. Thus, the above reasoning applies to VAE-like architectures.
\end{corollary}


\section{Computation experiment}
To validate the obtained theoretical results, experiments were conducted on the CIFAR-10 and FashionMNIST datasets. Below, we provide a detailed description of the methodology, model configurations, experimental results, and their connection to the proven theorems.

\subsection{Experimental settings}
For the experiments on CIFAR-10, the most distinct classes were selected. For the samples from each class, embeddings were computed using a pretrained ResNet50~\cite{ResNet}. The average distance between the vectors of different classes was then computed in the Euclidean space. The three classes with the largest pairwise distances were subsequently used in the experiment: 5 --- dogs, 6 --- frogs, and 8 --- ships.

The baseline models were an autoencoder (AE) and a variational autoencoder (VAE) with one and two fully connected layers in the encoder and decoder, both with and without a nonlinear activation function. The latent space had a dimension of $256$. All layers were initialized using the standard initialization and trained with the AdamW optimizer using a learning rate of $10^{-3}$ and a batch size of $32$. Each model was trained for $20$ epochs. To ensure the stability and statistical significance of the results, the experiments were repeated $N = 5$ times using different random subsamples. For a number of evaluations, the mean value and standard deviation are reported.

For each trained model, an embedding was computed using the algorithm described above. The constant $r$ from the algorithm was chosen such that the singular values retained in the vector accounted for $80\%$ of the sum of all singular values of the model parameters. Classification was performed using logistic regression with $L_2$ regularization, trained on the model embeddings. The target variable was the data class on which the corresponding autoencoder was trained. Classification performance was evaluated using Accuracy, Average Precision, and ROC AUC.

Hereafter, classes from the CIFAR-10 or FashionMNIST datasets are understood as two distinct datasets, each generated from its own distribution $p_c$. The class of a dataset $c$ refers to the image class defined in the original dataset.

\subsection{Baseline experiment (two classes)}
In the baseline experiment, we selected three most distinct classes from CIFAR-10. Subsamples of a fixed size $n$ were drawn from the first two classes. Here and below, $n=2000$ unless otherwise specified. An AE/VAE with the architecture described above was trained on each subsample; the models were then vectorized using the previously described algorithm, and logistic regression was trained on the resulting vectors for binary classification.

\begin{figure}[ht]
    \centering
    \includegraphics[scale=0.5]{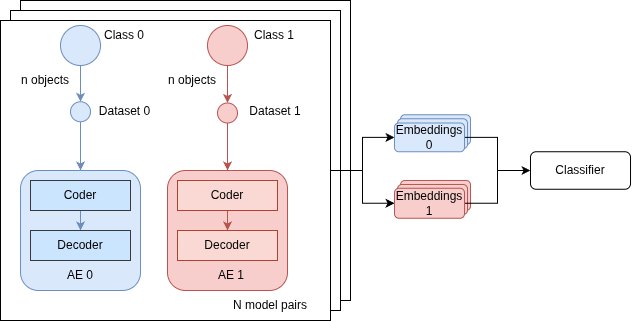}
    \caption{Experimental setup with training a set of autoencoders on subsamples from two classes followed by classification of their vector representations}
    \label{fg:model_1}
\end{figure}

An illustration of the experimental setup is shown in Fig.~\ref{fg:model_1}. The Accuracy of logistic regression in determining the class of the CIFAR-10 dataset was $0.98 (\pm 0.02)$ for an AE with two fully connected layers in the encoder and decoder and nonlinear activation functions; $0.97 (\pm 0.03)$ for a VAE; and $0.99 (\pm 0.01)$ for an AE with one fully connected layer in the encoder and decoder and without nonlinear activation functions.

\paragraph{Conclusion }
The logistic regression metrics for determining the class of the dataset on which the AE/VAE was trained are close to 1. Therefore, the singular value vector of the model contains sufficient information about the statistical properties of the training dataset. The experiment was conducted for models both with and without nonlinearities. The results were similar in both cases.

\subsection{Mixing third class experiment}
The next experiment examined the ability of model singular value vectors to reflect the mixing of training datasets. In the experiment, $\mathbf{X}_c$, ${c \in \{1, 2\}}$, were selected as sets of samples from two classes of CIFAR-10 or FashionMNIST, following the same procedure as in the previous experiment. A third class $\mathbf{X}_3$ was then mixed into these datasets with varying proportions $\alpha$. For a fixed set of model parameters, $\alpha$ was varied from $0.6$ to $0.99$, and new subsamples were generated for each value of $\alpha$, on which autoencoders were trained. The quality of binary classification was then evaluated using the resulting vector representations. Specifically, we evaluated the accuracy of logistic regression in determining the class ${c \in \{1, 2\}}$ on which the vectorized model was trained.

\begin{figure}[!tbph]
    \centering
    \includegraphics[width=\textwidth]{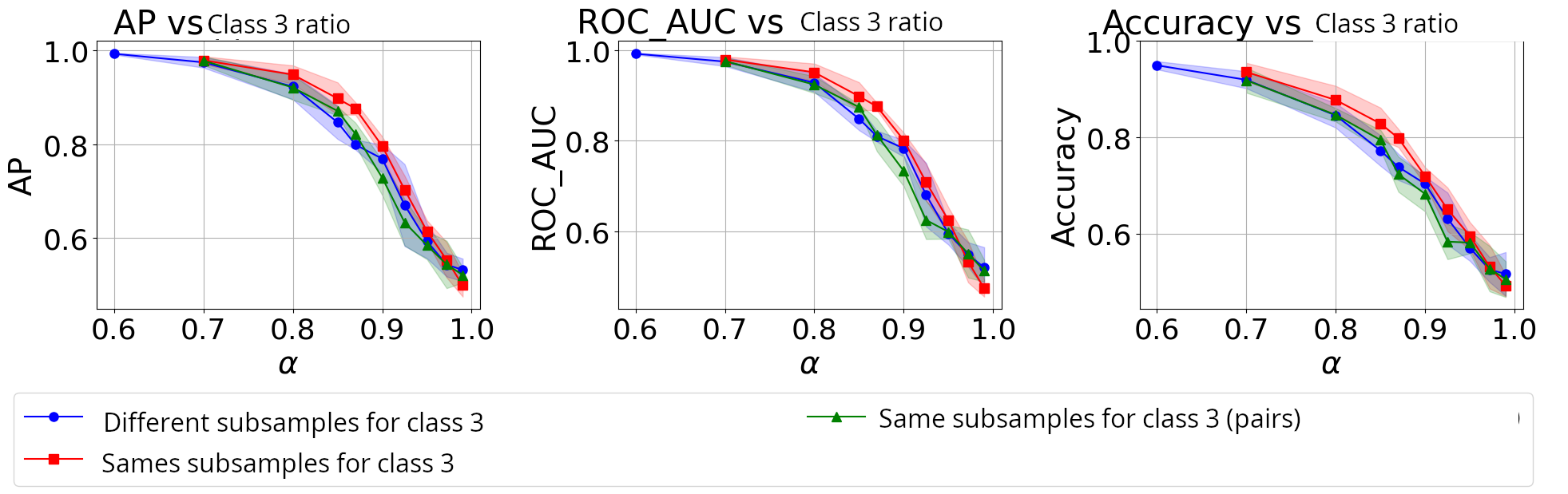}
    \par\medskip
    \textbf{(a)} CIFAR10, AE with 2 fully-connected layers in encoder and decoder, nonlinear activations
    \label{fg:exp_2a}

    \vspace{0.5cm}
    
    \includegraphics[width=\textwidth]{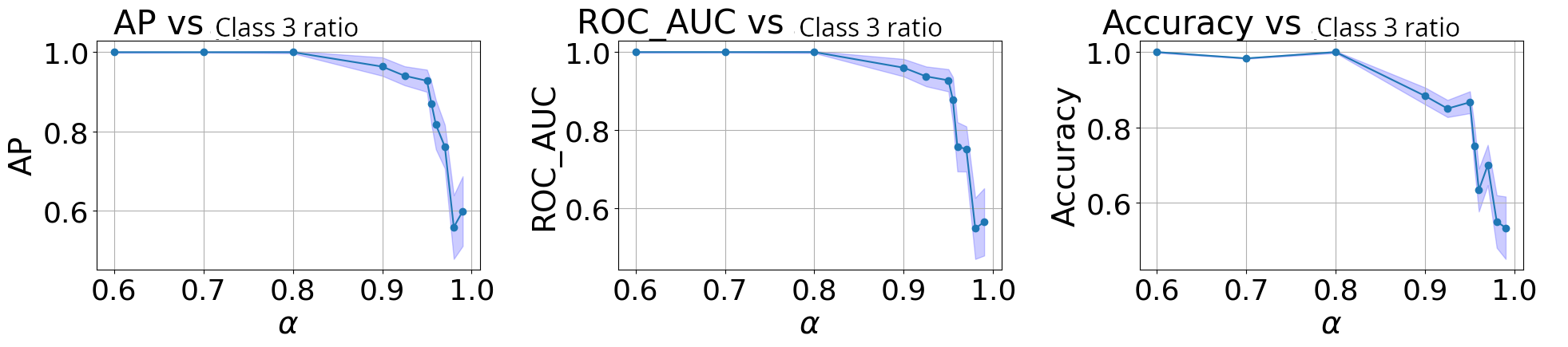}
    \par\medskip
    \textbf{(b)} IFAR10, AE with 1 fully-connected layer in encoder and decoder, without nonlinear activations (only for the case of different  subsamples for 3 classes)
    
    \label{fg:exp_2b}
    
    \vspace{0.5cm}
    
    \includegraphics[width=\textwidth]{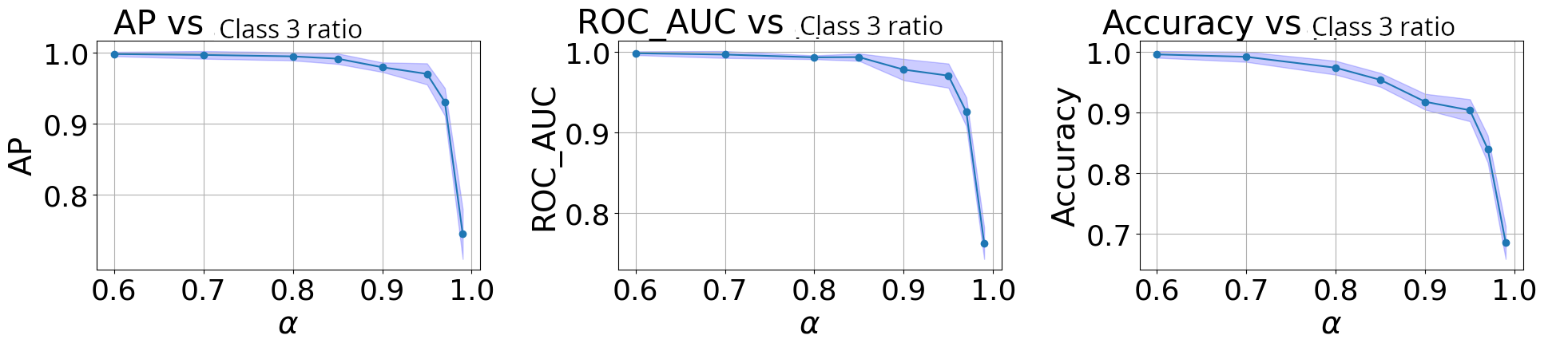}
    \par\medskip
    \textbf{(c)} FashionMNIST, AE with 2 fully-connected layer in encoder and decoder,  nonlinear activations (only for the case of different  subsamples for 3 classes)
    \label{fg:exp_2c}
    
    \caption{Demonstration of the performance of logistic regression on autoencoder embeddings for different proportions of the third class mixed into the models' training datasets}.
    \label{fg:exp_2}
\end{figure}

Fig.~\ref{fg:exp_2} shows the dependence of classification performance on the proportion of $\mathbf{X}_3$ in the training dataset. A sharp decrease in the metrics can be observed as $\alpha$ increases: a small number of samples from the third class (small $\alpha$) has almost no effect on separability, i.e., the performance metrics in Fig.~\ref{fg:exp_2} remain close to 1. In contrast, for large $\alpha$, the classifier of model embeddings begins to make more errors, as the distributions of the training data of the models become increasingly close to the distribution of $\mathbf{X}_3$, as schematically illustrated in Fig.~\ref{fg:classes}.

As an extension of the experiment, different sampling strategies were used for mixing the third class into $\mathbf{X}_1$ and $\mathbf{X}_2$. For each $\alpha$, a subsample was drawn independently and randomly for each of the $2N$ autoencoders (Different subsamples of class 3 in Fig.~\ref{fg:exp_2}); for each $\alpha$, the third-class subsample was drawn once and added identically to all $2N$ models (Same subsamples of class 3 (all) in Fig.~\ref{fg:exp_2}); in the third approach, for each $\alpha$, one of the $N$ autoencoders from each of the two main classes was trained on datasets with the same mixture of the third class (Same subsamples of class 3 (pairs) in Fig.~\ref{fg:exp_2}).

The experiment was conducted on two datasets: CIFAR-10 and FashionMNIST. We can observe a difference between the datasets in the proportion of the third class at which the performance of logistic regression in identifying the training dataset of a model begins to deteriorate substantially. This observation is explained in the following experiments and is related to the initial similarity between the classes selected for training.

\paragraph{Conclusion }
The proposed model vectorization method is sensitive to dataset mixing: at low proportions of the foreign class (small $\alpha$), the representations of models trained on subsamples from different classes remain highly separable, as shown in the plots in Fig.~\ref{fg:exp_2}. At larger $\alpha$, the representations of models trained on mixtures of $\mathbf{X}_c$ and $\mathbf{X}_3$ shift for $c \in \{1, 2\}$, and the classifier performs increasingly worse as $\alpha$ increases. This demonstrates that the embedding space reflects continuous transitions between data distributions. The experiment was conducted for autoencoders both without nonlinearities and with nonlinear activation functions. In all model variants, similar behavior was observed: as $\alpha$ increases, the accuracy of identifying the training dataset from the model embedding decreases.

\subsection{Random samples addition experiment}
In this experiment, $\mathbf{X}_3$ consists not of a single specific class from the CIFAR-10 dataset, but of any samples that do not belong to the classes from which $\mathbf{X}_c$, ${c \in \{1, 2\}}$, are selected. Thus, the previous experiment is extended to include a ``noise'' mixture. This type of noise simulates a real-world situation in which a dataset contains outliers or samples of unknown origin. The methodology follows the previous experiment: the proportion of noise samples was varied, new datasets $\tilde{\mathbf{X}}_1$ and $\tilde{\mathbf{X}}_2$ were sampled, models were trained on them, singular value vectors were constructed from the resulting models, and classification performance was evaluated using logistic regression.

\begin{figure}[ht]
    \centering
    \includegraphics[width=\linewidth]{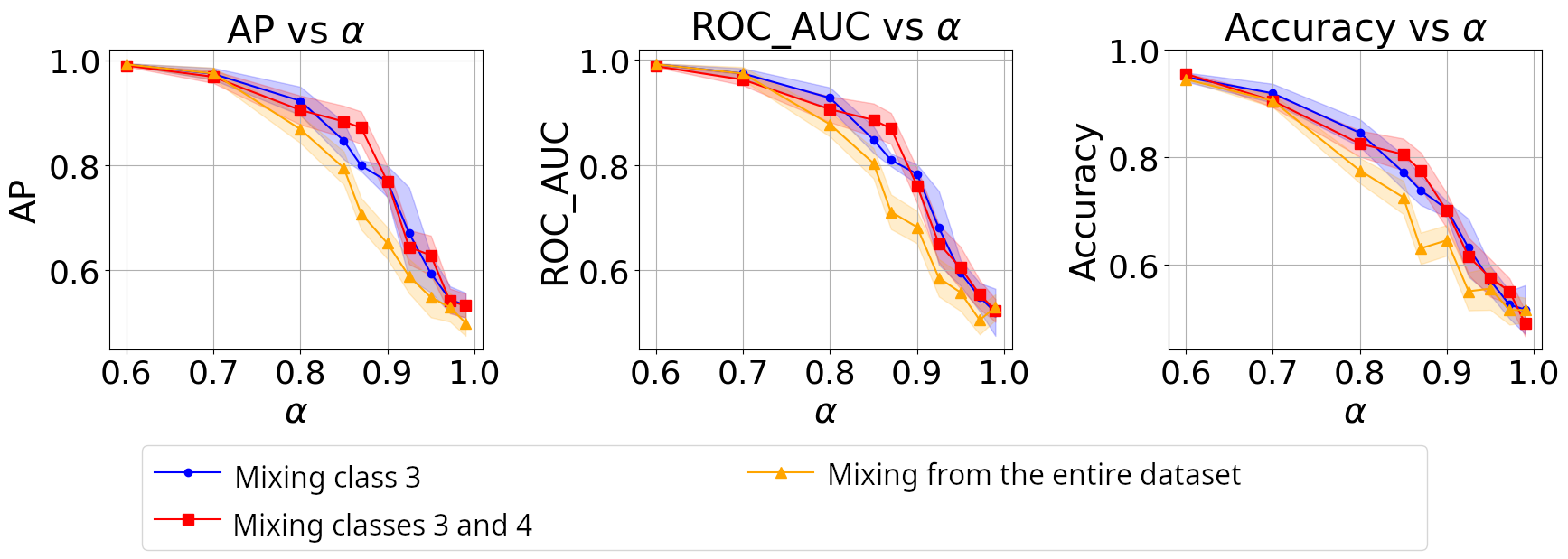}
    \caption{Demonstration of the effect of mixing different types of noise on model classification performance. Mixing only class 3: $\mathbf{X}_3$ is sampled from a single class; mixing classes 3 and 4: samples are drawn from two classes; mixing from the entire dataset: samples are drawn from all CIFAR-10 classes except those used in $\mathbf{X}_1$ and $\mathbf{X}_2$. }
    \label{fg:exp_3}
\end{figure}

\paragraph{Conclusion }  Adding random samples $\mathbf{X}_3$ reduces the separability (Fig.~\ref{fg:classes}) of the model embeddings trained on the mixtures $\tilde{\mathbf{X}}_1, \tilde{\mathbf{X}}_2$. Despite the small difference in the final accuracy, mixing noisy samples has a stronger effect on the separation performance, as shown by the yellow curve in Fig.~\ref{fg:exp_3}. In the previous experiment, vectors from only one specific class were mixed in. The results of this experiment support Lemma~\ref{lemma:1} and Theorem~\ref{theorem:1}. In particular, the greater the difference between the noise and $\tilde{\mathbf{X}}_1, \tilde{\mathbf{X}}_2$, the faster the performance deteriorates as the proportion of noise increases.

\subsection{Dataset singular values correlation}
This experiment analyzes the dependence of the convergence of the singular value vectors of model weights and the singular value vectors of the data on the proportion of samples from a new class mixed into the dataset. The theoretical convergence of the vectors was described for the special case of fully connected layers without nonlinearities in Theorem~\ref{theorem:1}. Three classes from the FashionMNIST dataset were selected. Samples from the first two classes were used to construct $\mathbf{X}_1$ and $\mathbf{X}_2$. The third class served as the mixed-in dataset $\mathbf{X}_3$. The resulting datasets $\tilde{\mathbf{X}}_1$ and $\tilde{\mathbf{X}}_2$ were obtained by sampling ${(1 - \alpha) n}$ samples from $\mathbf{X}_1$ and $\mathbf{X}_2$, respectively, and $\alpha n$ samples from $\mathbf{X}_3$, where ${n = 2000}$. Then, for values of $\alpha$ ranging from $0.6$ to $0.99$, the singular value vector was computed for each of the datasets $\tilde{\mathbf{X}}_1$ and $\tilde{\mathbf{X}}_2$. The mean absolute distance between the resulting vectors was computed over ${N = 100}$ different pairs of datasets $\tilde{\mathbf{X}}_1, \tilde{\mathbf{X}}_2$.

In the second part of the experiment, variational autoencoders were trained on $\tilde{\mathbf{X}}_1$ and $\tilde{\mathbf{X}}_2$, with ${N = 100}$ models trained for each class. The singular value vectors of the parameters obtained from the trained VAE models were then computed. For VAEs trained on datasets corresponding to different initial classes, the mean absolute distance between their embeddings was evaluated for different values of $\alpha$.

\begin{figure}[!ht]
    \centering
    \includegraphics[scale=0.35]{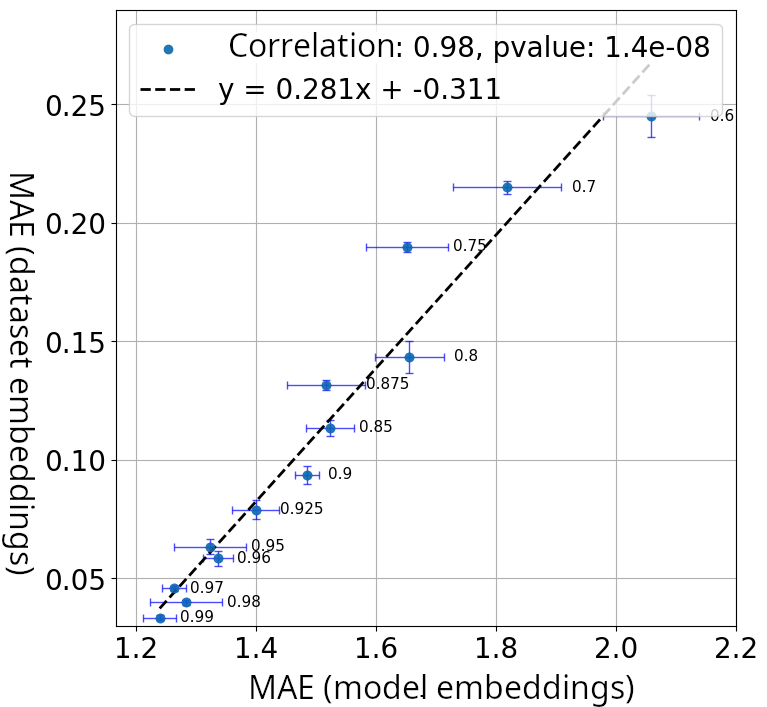}
    \caption{Comparison of the MAE between the data vectors and model embeddings. The X-axis shows the difference between the model embeddings, and the Y-axis shows the difference between the data vectors. Each point corresponds to a particular $\alpha$, the proportion of the third class in the training dataset.}
    \label{fg:exp_5}
\end{figure}

The mean absolute distance for the first and second parts of the experiment is shown in Fig.~\ref{fg:exp_5} along the $Y$-axis and $X$-axis, respectively.

\paragraph{conclusion } Fig.~\ref{fg:exp_5} shows a linear relationship between the mean absolute distance between the singular value vectors of the datasets and that between the singular value vectors of the models. According to a statistical test based on Pearson's correlation, the linear relationship is statistically significant.

\subsection{Pairwise comparison (One-vs-One)}
\begin{figure}[!ht]
    \centering
    \includegraphics[scale=0.35]{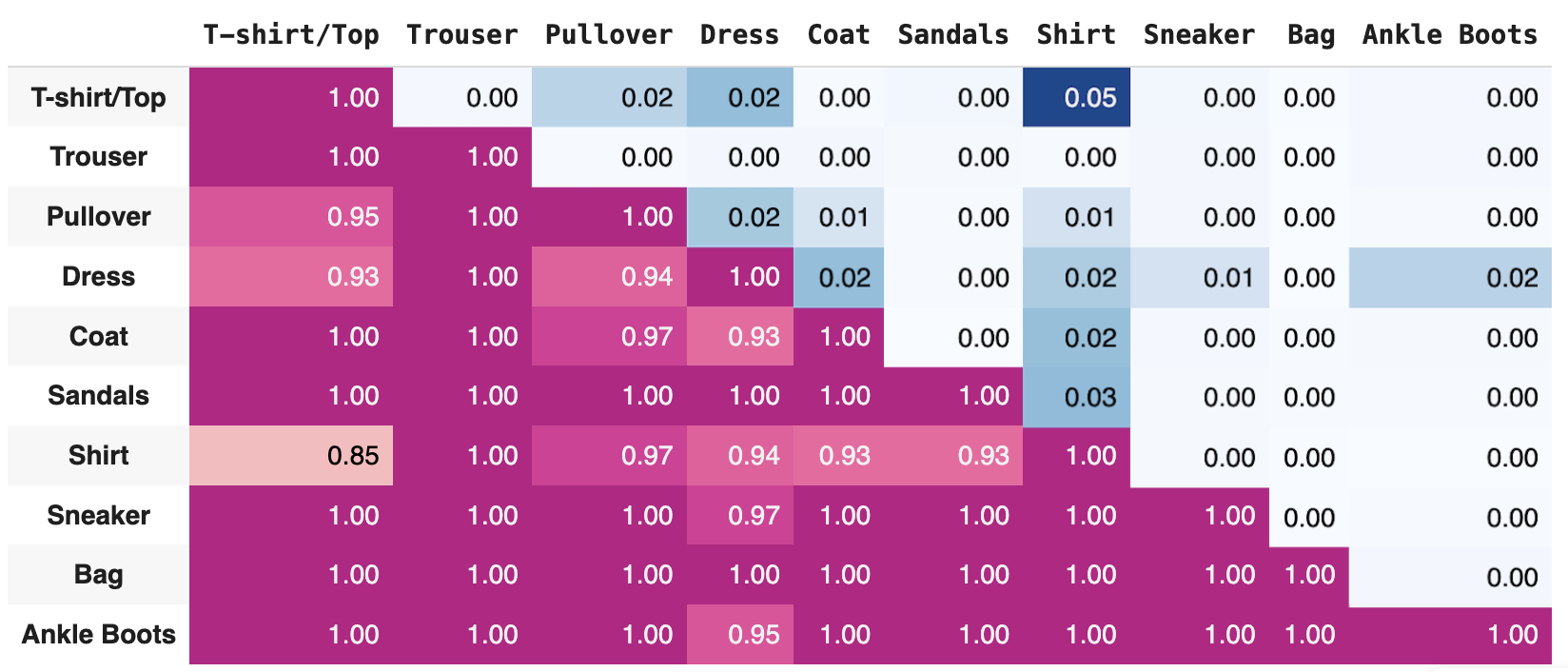}
    \caption{Demonstration of the effect of mixing different types of noise on model classification performance: the lower triangular matrix shows the mean performance of logistic regression in the model-vector separation experiment; the upper triangular matrix shows the mean deviation of logistic regression performance; the diagonal shows the mean performance of logistic regression in the one-vs-rest experiment.}
    \label{fg:exp_4}
\end{figure}

As an extension of the baseline experiment, a ``each class against each class'' comparison was performed. For all pairs of FashionMNIST classes, procedures analogous to those in the baseline experiment were carried out: subsamples $\mathbf{X}_1, \mathbf{X}_2$ without mixtures were constructed, a set of AEs with two fully connected layers and nonlinear activation functions was trained, the models were vectorized, and a binary classifier was trained. The results were averaged and visualized as a pairwise separability matrix in Fig.~\ref{fg:exp_4}.

\paragraph{Conclusion } The resulting pairwise matrix shows a wide range of separability: some class pairs (e.g., ``coat'' versus ``dress'') yield nearly perfect classification based on the singular value vectors of the models. Other pairs, such as visually similar classes, are more difficult to distinguish. Consequently, the logistic regression performance for these pairs is comparatively low. These observations are consistent with expectations: the greater the difference between the statistics of the original classes, the more pronounced the signal in the singular spectra of the autoencoder parameters.

\subsection{Experiment discussion}
Across all series of experiments, the key hypothesis was confirmed: the singular value vector of the autoencoder parameters contains informative signals about the distribution of the training data, both with and without a nonlinear activation function. For statistically distinct classes, a classifier trained on such vectors achieves nearly perfect separability, which is consistent with the conclusions of Theorem~\ref{thm:vae_singulars}. When classes are mixed and noise is added, a predictable decrease in performance is observed, indicating that the embedding space correctly reflects statistical shifts in the data.


\section{Conclusion}
This work has shown that even when using a simple autoencoder architecture---one or two fully connected layers in the encoder and decoder---and a minimalist approach to feature representation based on singular value vectors, the parameters of trained models allow for nearly perfect discrimination between datasets with significant differences in the singular value spectra of their covariance matrices. This result is particularly noteworthy because it is achieved without the use of complex architectures, pretrained models, or heuristic methods for computing statistics of the input data.

Thus, a trained neural network can be viewed as a universal means of describing a dataset with a single vector formed exclusively from its parameters. Unlike traditional approaches based on specially selected statistical features or data histograms, the informative representation here emerges as a by-product of the training process.

The conducted experiments---from pairwise class comparisons to the analysis of mixtures and the addition of noisy samples---demonstrated that the space of such vectors adequately reflects the degree of difference between datasets and is sensitive to their perturbations. This opens up the possibility of applying the approach to tasks such as dataset retrieval and comparison, distribution shift analysis, and automatic categorization, without the need to store or directly access the data themselves.

A promising direction for future work is to investigate the observed effect using other approaches and to analyze their performance across different architectures with varying depth and complexity.

\newpage

\end{document}